\documentclass[letterpaper]{article} 
\usepackage{aaai2026}  
\usepackage{times}  
\usepackage{helvet}  
\usepackage{courier}  
\usepackage[hyphens]{url}  
\usepackage{graphicx} 
\usepackage{natbib}  
\usepackage{caption} 
\usepackage{algorithm}
\usepackage{algorithmic}
\usepackage[utf8]{inputenc} 
\usepackage[T1]{fontenc}    
\usepackage{url}            
\usepackage{booktabs}       
\usepackage{amsfonts}       
\usepackage{nicefrac}       
\usepackage{microtype}      
\usepackage{amsmath}
\usepackage{graphicx} 
\usepackage{subcaption} 
\usepackage{tikz} 
\usepackage{caption} 
\usepackage{float} 
\usepackage{multirow} 
\usepackage{multicol} 
\definecolor{highlight}{HTML}{FEEEE1}  
\usepackage{booktabs} 
\usepackage{tcolorbox} 
\tcbuselibrary{listings,breakable}
\usepackage{newfloat}
\usepackage{listings}

\usepackage{algorithm}
\usepackage{algorithmic}

\usepackage{newfloat}
\usepackage{listings}
\DeclareCaptionStyle{ruled}{labelfont=normalfont,labelsep=colon,strut=off} 
\floatstyle{ruled}
\newfloat{listing}{tb}{lst}{}
\floatname{listing}{Listing}
\title{MuGa-VTON: Multi-Garment Virtual Try-On via Diffusion Transformers with Prompt Customization}

\author{
    Ankan Deria\textsuperscript{1, 4},
    Dwarikanath Mahapatra\textsuperscript{2},
    Behzad Bozorgtabar\textsuperscript{3},
    Mohna Chakraborty\textsuperscript{4},
    Snehashis Chakraborty\textsuperscript{4},
    Sudipta Roy\textsuperscript{4, *} \\  
}
\affiliations{
    \textsuperscript{1} Mohamed bin Zayed University of Artificial Intelligence, UAE \\
    \textsuperscript{2} Khalifa University, UAE \\
    \textsuperscript{3} École Polytechnique Fédérale de Lausanne (EPFL), Switzerland \\
    \textsuperscript{4} Jio Institute, India 
    
}

\usepackage{bibentry}

\begin{document}

\twocolumn[{%
\renewcommand\twocolumn[1][]{#1}%
\maketitle
\begin{center}
    \centering
    \includegraphics[width=1.0\textwidth]{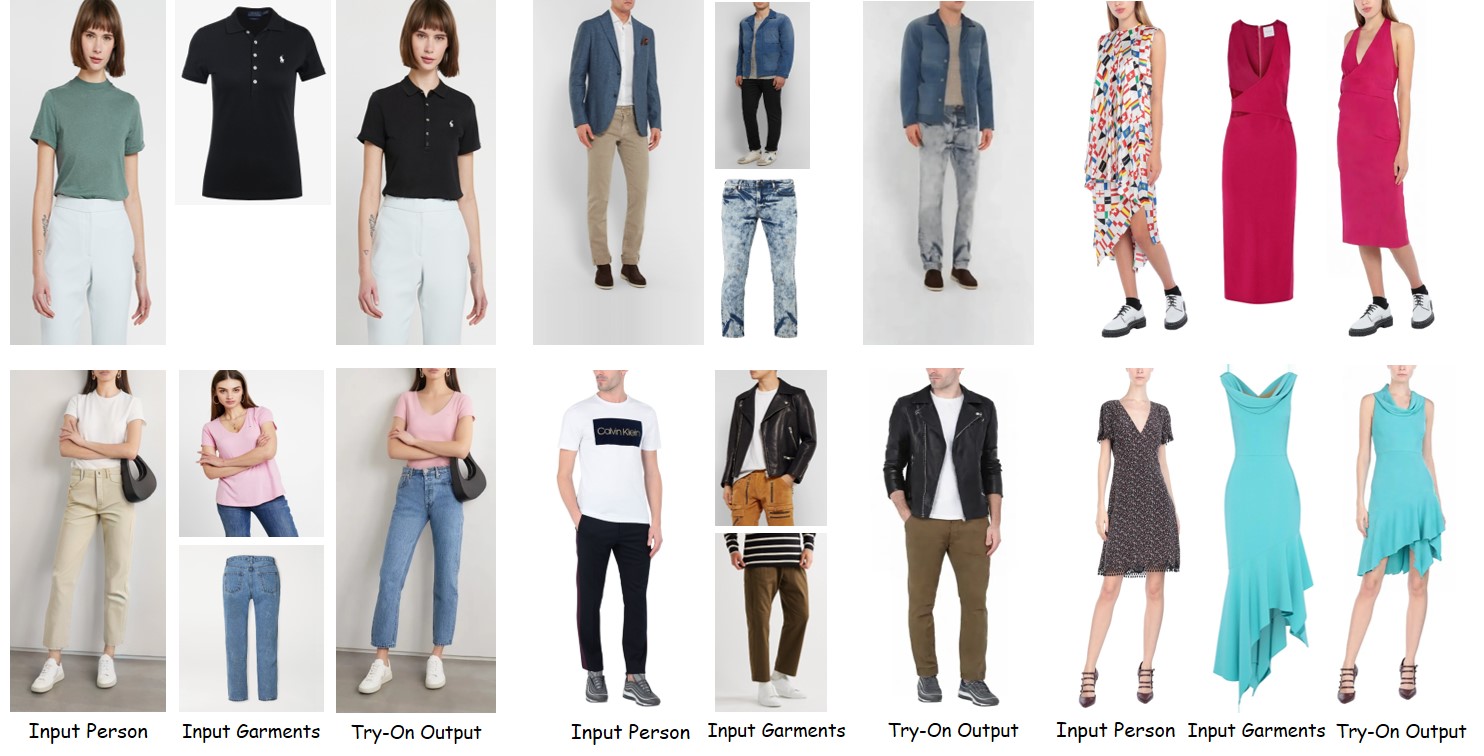}
    \vspace{-5pt}
    \captionof{figure}{\small  
        MuGa-VTON generates realistic virtual try-on results given a single person image and multiple garment inputs. The model preserves distinctive personal attributes such as hand tattoos (top right), accessories like handbags (bottom left), and supports prompt-based edits, e.g., rolling up the outer jacket sleeves (bottom right).
    }
    \label{fig:intro_fig}
\end{center}%
}]

\begin{abstract}

Virtual try‑on seeks to generate photorealistic images of individuals in desired garments, a task that must simultaneously preserve personal identity and garment fidelity for practical use in fashion retail and personalization. However, existing methods typically handle upper and lower garments separately, rely on heavy preprocessing, and often fail to preserve person-specific cues such as tattoos, accessories, and body shape-resulting in limited realism and flexibility.  
To this end, we introduce \textbf{MuGa‑VTON}, a unified multi-garment diffusion framework that jointly models upper and lower garments together with person identity in a shared latent space. Specifically, we proposed three key modules: the \emph{Garment Representation Module (GRM)} for capturing both garment semantics, the \emph{Person Representation Module (PRM)} for encoding identity and pose cues, and the \emph{A‑DiT} fusion module, which integrates garment, person, and text‑prompt features through a diffusion transformer. This architecture supports prompt‑based customization, allowing fine‑grained garment modifications with minimal user input.  
Extensive experiments on the VITON-HD and DressCode benchmarks demonstrate that MuGa-VTON outperforms existing methods in both qualitative and quantitative evaluations, producing high-fidelity, identity-preserving results suitable for real-world virtual try-on applications.

\end{abstract}


\section{Introduction}
\label{sec:intro}


As virtual try-on systems become adopted in applications such as online fashion retail, users expect personalized results with minimal input. Modern consumers seek not only garment swapping but also fine-grained customization, such as preserving tattoos or accessories and adjusting garment styles, without requiring extensive manual preprocessing.

While recent works \cite{kim2024stableviton, choi2021viton, yang2024texture, choi2024improving, liang2024vton, zeng2024cat} have shown promising results, they remain limited to single-garment settings and rely heavily on pre-processing such as agnostic masks, keypoints, segmentation maps, and DensePose. Despite such rich input representations, these studies often fail to preserve identity-specific attributes, including tattoos, accessories, scars, or detailed body shape, which are critical for realism and user satisfaction. Moreover, their inability to generate coordinated upper and lower garments prevents the creation of stylistically coherent outfits, limiting their applicability in real‑world scenarios.

Achieving high-fidelity virtual try-on introduces two core technical challenges: generating accurate garment masks and preserving high-frequency details when modeling multiple garments simultaneously. Existing datasets such as VITON-HD \cite{choi2021viton} and DeepFashion2 \cite{ge2019deepfashion2} rely on coarse agnostic masks that discard subtle appearance details, often causing artifacts or incomplete garment transfer. To address this, we employ \textbf{Sapiens} \cite{khirodkar2024sapiens}, a high-precision human body-part segmentation model, to produce refined masks that use mask the garment precisely in training for better visible identity-specific attributes (additional details are provided in the supplementary material).  
Furthermore, we leverage GPT‑4o to generate structured garment descriptions that serve as rich semantic conditioning for prompt-driven training, enabling fine-grained garment editing and customization. To manage the complexity of modeling multiple garments without losing detail, we adopt a progressive training strategy inspired by M\&M VTO \cite{zhu2024m}, starting with low-resolution inputs and gradually increasing resolution so the model first learns coarse structure and subsequently refines intricate details.

\begin{figure*}
    \centering
    \includegraphics[width=0.99\linewidth]{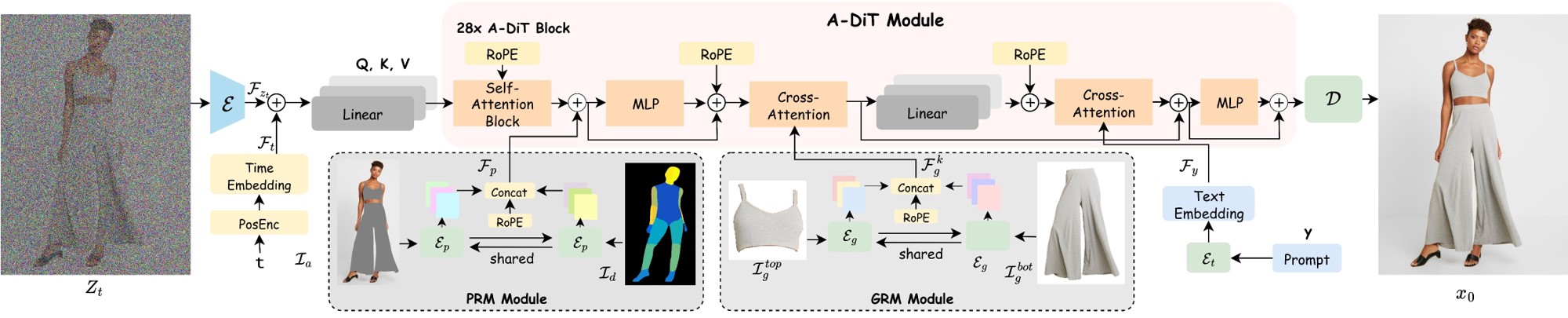}
    \caption{
        \textbf{MuGa‑VTON architecture.} 
        VAE encoders ($\mathcal{E}$, $\mathcal{E}_p$, $\mathcal{E}_g$) extract feature maps ($\mathcal{F}_{z_t}$, $\mathcal{F}_p$, $\mathcal{F}_g^{k}$) from input images. The diffusion timestep $t$ and text prompt $y$ are embedded as conditioning tokens $(\mathcal{F}_t, \mathcal{F}_y)$. 
        (a) \emph{PRM module:} Person image $\mathcal{I}_a$ and DensePose map $\mathcal{I}_d$ are encoded by $\mathcal{E}_p$ to produce person feature tokens $\mathcal{F}_p$. 
        (b) \emph{GRM module:} Garment images $\mathcal{I}_g^{k}$ with $k \in \{\text{upper}, \text{lower}, \text{full}\}$ are encoded by $\mathcal{E}_g$ to generate garment tokens $\mathcal{F}_g^{k}$. 
        (c) \emph{A‑DiT module:} Serves as the central fusion block where person tokens $\mathcal{F}_p$, garment tokens $\mathcal{F}_g^{k}$, and text‑prompt tokens $\mathcal{F}_y$ interact. RoPE preserves spatial relationships across resolutions, while the resulting tokens modulate self‑attention and cross‑attention layers to align identity with garment semantics for realistic synthesis.
        The aligned tokens are decoded by $\mathcal{D}$, a decoder symmetric to $\mathcal{E}$, to reconstruct the clean latent $x_0$.
    }
    \label{fig:A-DiT Architecture}
    \vspace{-5pt}
\end{figure*}

In this work, we present \textbf{MuGa-VTON}, a unified framework for \textbf{Mu}lti-\textbf{Ga}rment \textbf{VI}sual-\textbf{T}ry-\textbf{O}n and interactive editing. The framework is built around three core components: the \emph{Person Representation Module (PRM)}, the \emph{Garment Representation Module (GRM)} and the \emph{A-DiT} fusion module in the Figure~\ref{fig:A-DiT Architecture}. PRM encodes visible person features and DensePose maps, integrating them with Rotary Positional Embedding (RoPE) to obtain identity and pose-aware tokens. GRM similarly encodes upper and lower garments with RoPE to capture rich garment semantics and structure. These representations are then fused within A-DiT, a diffusion transformer that jointly aligns person and garment tokens with optional text-prompt embeddings to generate fully customized try-on results. This design supports both standalone product images and garments cropped from real-world photos, combined with a single full-body image of the target person. Text instructions, such as “tuck in the shirt” or “roll up the sleeves”, further enable fine-grained style control. By faithfully preserving garment fidelity and identity-specific attributes, MuGa-VTON produces highly realistic and personalized try-on results, as illustrated in Figure~\ref{fig:intro_fig}.

Extensive experiments demonstrate that MuGa‑VTON achieves state-of-the-art performance across quantitative metrics and user preference studies. Our results highlight substantial improvements in garment fidelity, identity preservation, and customization, establishing MuGa‑VTON as a practical solution for real-world VTO applications requiring high realism and personalization.\\


In summary, our main contributions are as follows: 
\begin{itemize}
    \item We propose MuGa‑VTON, a unified framework that jointly models upper and lower garments in a shared latent space and supports prompt‑based customization for flexible outfit generation.  
    \item Our design preserves fine‑grained identity attributes, including tattoos, scars, muscle tone and accessories, enabling highly realistic and personalized VTO results.  
    \item MuGa‑VTON requires minimal user input and avoids extensive preprocessing, making it practical for real‑world fashion and e‑commerce deployment.  
    \item We adopt a progressive training strategy that first captures coarse garment structure and then refines high‑frequency details, achieving high fidelity without requiring super‑resolution stages. 
\end{itemize}


\section{Related Work}
\label{sec:related_work}

\subsection{Virtual Try-On Approaches}

Early stage virtual try-on works \cite{choi2021viton,xie2023gp,yang2023occlumix} introduced multi-stage GAN-based pipelines for garment alignment and person reconstruction. Although these studies utilized numerous input features, they often failed to preserve fine garment details and ensure a proper garment fit. For instance, HR-VITON \cite{lee2022high} partially addressed alignment issues by generating intermediate target cloths, yet struggled to accurately replicate muscle structure, body shape, and skin tone, limiting the overall realism of the synthesized outputs. 

Building on these efforts, StableVITON \cite{kim2024stableviton} leveraged a pre-trained diffusion model with zero cross-attention blocks to capture semantic correspondences, thereby eliminating the need for independent warping. Post-processing solutions such as Handrefiner \cite{lu2024handrefiner} and RHanDS \cite{wang2024rhands} further refined malformed hand reconstructions using hand mesh models. In parallel, VTON-HandFit \cite{liang2024vton} shifted focus to accurately reconstructing hands and fingers by incorporating hand priors and specialized loss functions.
Other studies \cite{choi2024improving, shen2024imagdressing} have sought to integrate high-level and low-level features more effectively. For example, IDM-VTON \cite{choi2024improving} employed advanced attention modules to capture both garment semantics and fine-grained details, while IMAGDressing-v1 \cite{shen2024imagdressing} used dual UNet networks with hybrid attention mechanisms and image captions to enhance garment detail generation. More recent studies such as LADI-VTON \cite{morelli2023ladi} and DCI-VTON \cite{gou2023taming} have explored treating clothing as pseudo-words or incorporating warping networks into pre-trained diffusion models. Street TryOn \cite{cui2024street} tackled the challenge of outdoor virtual try-on by integrating garment edge masks with inpainting techniques to better blend garments with complex backgrounds, utilizing filtered images from the DeepFashion2 \cite{ge2019deepfashion2} and VITON-HD \cite{choi2021viton} datasets. To further deal with multi-part references, Parts2Whole \cite{huang2024parts} employed a semantic-aware encoder with mask-guided shared self-attention. However, despite improved garment composition, this approach often yielded overly smooth, cartoon-like outputs and incurred increased inference times.


Further advancing VTO studies, TryOnDiffusion \cite{zhu2023tryondiffusion} proposed a Parallel-UNet architecture that enabled implicit garment warping and seamless blending in a single pass, managing large occlusions and pose variations with a cascaded diffusion model for high-resolution outputs. More recently, M\&M VTO \cite{zhu2024m} introduced a single-stage diffusion model generating high-quality images by disentangling person-specific and garment-specific features, enabling efficient fine-tuning for identity preservation. However, they rely on synthetic data for fine-tuning, which can introduce biases and limit their generalizability.

Despite these advancements, existing models still struggle to generate outputs that fully preserve all distinguishing identity attributes while remaining flexible for multi‑garment settings. Most approaches are restricted to single‑garment visualization and depend on extensive auxiliary inputs. In this work, we present \textbf{MuGa‑VTON}, a unified framework that jointly handles upper and lower garments, preserves person identity, and supports prompt‑based customization for fine‑grained editing.

\subsection{Image Editing with Diffusion Models}
Diffusion-based image editing techniques initially relied on using image masks or manipulating noise levels conditioned on text prompts. For instance, SDEdit \cite{meng2021sdedit} introduced a stochastic process that added noise to input images and progressively denoised them for editing. Similarly, BlendedDiffusion \cite{avrahami2022blended}, inspired by CLIP-guided diffusion \cite{crowson2021clip}, utilized the CLIP text encoder \cite{radford2021learning} along with spatial masks to blend noisy input images with locally generated content for localized edits. However, mask-based approaches are unsuitable for virtual try-on tasks, where precise edits, such as ``rolling up a shirt" are required.
To address these limitations, Prompt-to-Prompt (P2P) \cite{hertz2022prompt} enables text-driven image editing by adjusting cross-attention scores based on inverted latent representations. Methods like InstructPix2Pix \cite{brooks2023instructpix2pix} and Forgedit \cite{zhang2023forgedit} take a direct approach by manipulating images during the denoising process. These models, built on finetuned versions of Stable Diffusion, are trained on paired examples with specific editing instructions to achieve precise modifications.
It is crucial to handle language-based image editing techniques with great care to ensure that the original garment structure is preserved. The prompt should be injected strategically into the noise prediction module to avoid distorting the garment details. Our approach adopts a controlled prompt injection mechanism \cite{hertz2022prompt, deria2024inverge} within the A‑DiT module, enabling precise local garment edits, such as rolling sleeves or tucking shirts, while maintaining the structural integrity and texture continuity of both person and garment.


\section{Methodology}
\label{sec:methodology}

We present a single‑stage diffusion framework that synthesizes virtual try-on images at multiple resolutions during both training and inference. Unlike UNet‑based pipelines \cite{zhu2023tryondiffusion} that require additional super‑resolution stages, our DiT backbone captures global structure and fine‑grained details in a unified pass, enabling prompt‑based editing while preserving garment fidelity and personal identity. The framework consists of three key components: the \textbf{Person Representation Module (PRM)} for identity and pose encoding, the \textbf{Garment Representation Module (GRM)} for multi‑garment semantics, and the \textbf{A‑DiT} module that fuses these representations with text prompts to guide synthesis.

\begin{figure}
  \centering
      \includegraphics[width=\linewidth]{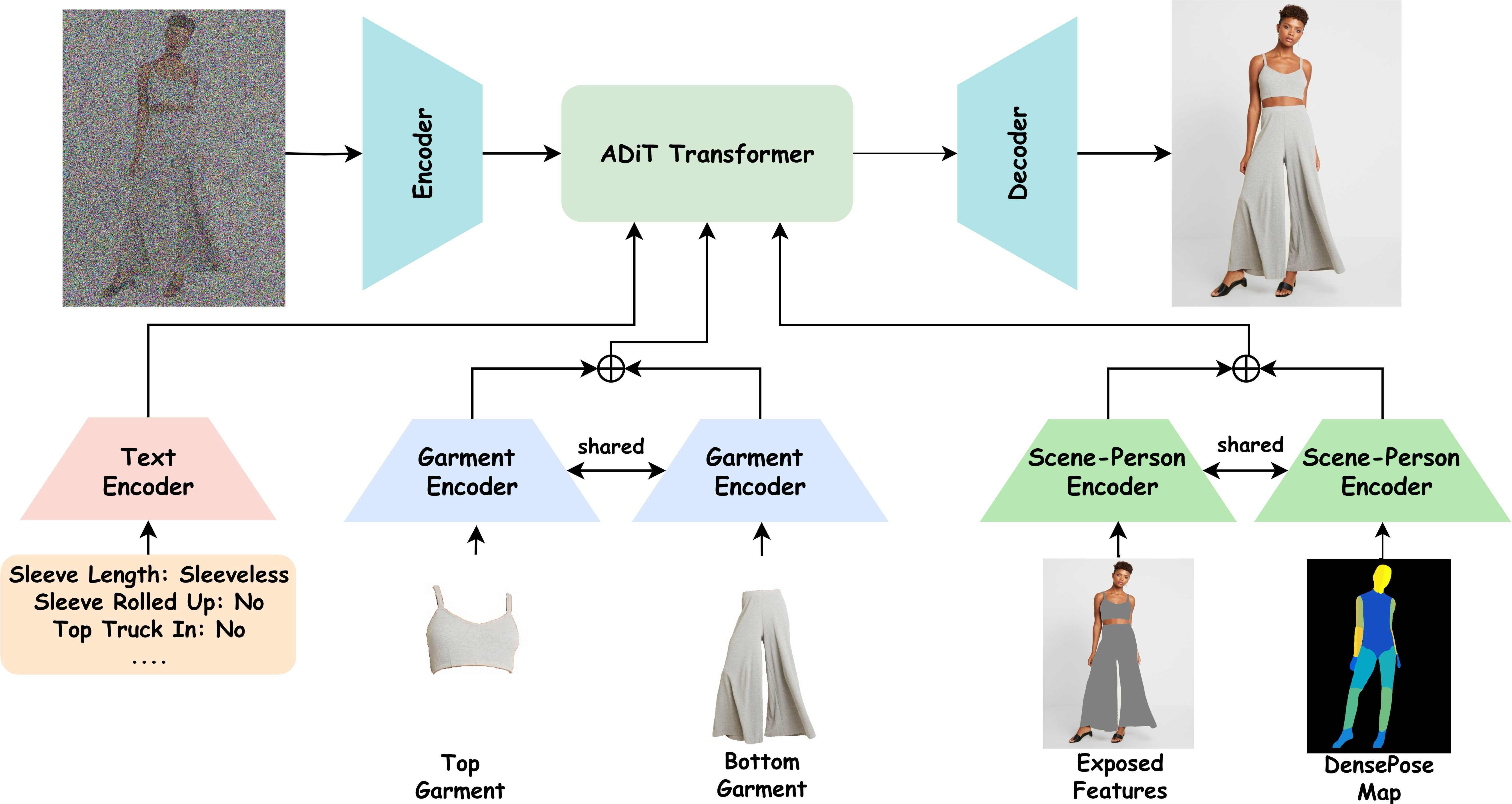}
      \caption{
          \textbf{Overview of MuGa‑VTON.} The model integrates multiple encoders and attribute modules to handle multi‑garment try‑on. It takes as input a person image, layout description, DensePose map, and garment images for both upper and lower clothing (with missing items replaced by blank images), enabling flexible virtual try‑on.
        }
      \label{fig: main_arct}
\end{figure}


\subsection{Dataset Preparation and Preprocessing}
\label{sec:data_process}

MuGa-VTON is trained using person images paired with corresponding upper and lower-garment images. The garment inputs may appear either as layflat product images or as garments worn by individuals. While widely adopted, the VITON-HD dataset \cite{choi2021viton} has notable shortcomings: it primarily provides layflat upper garments and employs coarse agnostic masks that discard important identity-specific features such as tattoos, scars, muscle tone, and accessories. As a result, residual parts of the original garment often remain visible, leading to incomplete clothing transfer.

To address these limitations, we generate refined masks using \textbf{Sapiens} \cite{khirodkar2024sapiens}, a high-precision human body-part segmentation model capable of segmenting 28 categories even in challenging poses. These masks enable accurate garment extraction and construction of improved agnostic masked-image $\mathcal{I}_a$. Furthermore, to balance the benefits of layflat and cropped garment representations, we augment the dataset to include both during training, enhancing robustness across diverse garment presentations.
For language-guided garment customization, we focus on realistic attribute edits such as rolling up sleeves or tucking in shirts rather than altering intrinsic properties like color or fabric texture. To support this, we employ GPT-4o-mini to generate structured garment descriptions that provide rich semantic cues for prompt-based training. Additional preprocessing details are provided in the supplementary material.

\subsection{Encoder}
\label{sec:encoder}

High‑fidelity multi‑garment try‑on requires a latent representation that preserves fine‑grained identity and garment semantics while remaining efficient for diffusion. Direct pixel‑space modeling is memory‑intensive and often blurs garment boundaries. To overcome this, A‑DiT uses an encoder $\mathcal{E}$ to compress person and garment inputs into a shared latent space for transformer‑based diffusion.

Formally, an input image $\mathbf{I} \in \mathbb{R}^{H \times W \times 3}$ is encoded by a pre‑trained Variational Autoencoder (VAE) into
\begin{equation}
    \mathbf{z} = \mathcal{E}(\mathbf{I}), \qquad \mathbf{z} \in \mathbb{R}^{c \times h \times w}, \; (h,w)\ll(H,W),
\end{equation}
where $\mathbf{z}$ preserves semantic structure at reduced spatial resolution, enabling efficient modeling within the diffusion process.

For text‑driven customization, the conditioning prompt $y$ is embedded via the CLIP text encoder \cite{radford2021learning}:
$\mathcal{F}_y = \mathcal{E}_{\text{t}}(y)$,
which ensures garment semantics and person attributes are aligned in the same latent space.

In practice, we adopt the high‑resolution VAE from SDXL \cite{podell2023sdxl}, fine‑tuned on $512\times512$ images, which improves upon SD‑1.5 \cite{rombach2022high} by producing sharper garment boundaries, reducing saturation artifacts, and preserving attributes such as tattoos and body contours.

\subsection{Person Representation Module(PRM)}
\label{sec:PRM}
The PRM module encodes identity and pose information from the target person. A shared scene-person encoder $\mathcal{E}_p$ processes the person agnostic masked-image $\mathcal{I}_a$ together with its DensePose map $\mathcal{I}_d$, producing multi‑scale feature tokens $\mathcal{F}_p$. While $\mathcal{I}_a$ preserves appearance cues such as tattoos, skin tone and accessories, $\mathcal{I}_d$ provides structural guidance crucial for aligning loose or complex garments. The extracted features are fused using RoPE and supplied to the A‑DiT module, where they interact with garment tokens from GRM through cross‑attention to achieve fine‑grained alignment and preserve high‑frequency identity details.

\subsection{Garment Representation Module(GRM)}
\label{sec:GRM}
The GRM module encodes garment inputs to provide semantic features for alignment with person features. A shared garment encoder $\mathcal{E}_g$ processes garment images $\mathcal{I}_g^{k}$, where $k \in \{\text{upper}, \text{lower}, \text{full}\}$, producing feature tokens $\mathcal{F}_g^{k}$ that capture texture and shape information. These tokens are embedded using RoPE to maintain spatial consistency across resolutions and are passed to the A‑DiT module, where they interact with person tokens $\mathcal{F}_p$ via cross‑attention for fine‑grained garment‑person fusion. This design supports precise garment integration and prompt‑based customization without requiring explicit warping.

\subsection{A-DiT Module}
\label{sec: A-DiT Module}

Multi‑garment try‑on requires fusing person, garment, and text‑prompt features while preserving spatial relationships and fine‑grained identity cues. The A‑DiT module performs this fusion directly in the latent space produced by the VAE, using  diffusion transformer.

Person features $\mathcal{F}_p$ from PRM are concatenated into the input image token stream $\mathcal{F}_{z_t}$, while garment features $\mathcal{F}_g^{k}$ from GRM are injected via cross‑attention. RoPE Embedding ensures spatial consistency and conditioning with text‑prompt embeddings $\mathcal{F}_y$ enables semantic garment‑person alignment.

Training follows the $v$‑prediction objective \cite{salimans2022progressive}, where the network predicts the velocity between noisy and clean latents instead of directly regressing the clean latent. This approach stabilizes high‑resolution synthesis and reduces color drift. Formally, the try‑on network is defined as
\begin{equation}
    {\mathbf{z}}_0 = \mathbf{x}_\theta(\mathbf{z}_t, t, \mathbf{c}_{\text{cond}}),
\end{equation}
where $t$ is the diffusion timestep, $\mathbf{z}_t$ is the noisy latent obtained by corrupting the ground‑truth $\mathbf{z}_0$, and $\mathbf{c}_{\text{cond}}$ represents fused PRM, GRM, and text‑prompt embeddings. Given the predicted velocity $\hat{\mathbf{v}}_t$, the clean latent is reconstructed as
\begin{equation}
    {\mathbf{z}}_0 = \delta_t \mathbf{z}_t - \eta_t \hat{\mathbf{v}}_t,
\end{equation} 
with $\delta_t,\eta_t \in (0,1)$ controlling the signal‑to‑noise ratio. 

\begin{table*}[t]
\centering
\resizebox{0.8\textwidth}{!}{%
    \begin{tabular}{l|cccc|cccc} 
    \toprule
    \multirow{2}{*}{\textbf{Method}}
        & \multicolumn{4}{c|}{\textbf{VITON-HD}}
        & \multicolumn{4}{c}{\textbf{DressCode Upper}} \\ 
    \cmidrule(r){2-5} \cmidrule(l){6-9}                
        & \textbf{SSIM $\uparrow$} & \textbf{LPIPS $\downarrow$} 
        & \textbf{FID $\downarrow$} & \textbf{KID $\downarrow$}
        & \textbf{SSIM $\uparrow$} & \textbf{LPIPS $\downarrow$}
        & \textbf{FID $\downarrow$} & \textbf{KID $\downarrow$} \\
    \midrule
    VITON-HD~\cite{choi2021viton}         & 0.856 & 0.119 & 12.564 & 3.26  &  --   &   --  &   --   &   --  \\
    HR-VITON~\cite{lee2022high}           & 0.868 & 0.106 & 11.785 & 2.82  & 0.904 & 0.069 & 13.84  & 2.92 \\
    GP-VTON~\cite{xie2023gp}              & \underline{0.891} & 0.105 &  \underline{6.430} &  --   & 0.913 & \underline{0.056} & 12.48  &  --  \\
    LADI-VTON~\cite{morelli2023ladi}      & 0.859 & 0.099 &  9.630 & 1.97  & 0.901 & 0.076 & 14.88  & 3.37 \\
    DCI-VTON~\cite{gou2023taming}         & 0.862 & 0.087 &  8.945 & 1.09  & 0.907 & 0.073 & 12.06  & 1.81 \\
    StableVITON~\cite{kim2024stableviton} & 0.881 & \underline{0.081} &  8.588 & 0.83  & 0.909 & 0.066 & 10.10  & \underline{0.96} \\
    CatVTON~\cite{chong2024catvton}       & 0.871 & 0.082 &  8.653 & 1.09  & 0.918 & 0.061 &  9.95  & 1.40 \\
    IDM--VTON~\cite{choi2024improving}    & 0.870 & 0.102 & \textbf{6.290} &  --  & 0.920 & 0.062 & \textbf{8.64} &  --  \\
    Any2AnyTryon~\cite{guo2025any2anytryon}  & 0.839 & 0.088 &  6.934 & \underline{0.74} & 0.8832 & 0.095 &  10.52 & 1.24 \\
    \textbf{Ours}                         & \textbf{0.898} & \textbf{0.073} & 7.240 & \textbf{0.52} & \textbf{0.937} & \textbf{0.046} & \underline{8.93} & \textbf{0.94} \\
    \bottomrule
    \end{tabular}
}

\caption{
Evaluation results on VITON‑HD and DressCode Upper datasets using four standard metrics: SSIM (Structural Similarity Index Measure), LPIPS (Learned Perceptual Image Patch Similarity), FID (Fréchet Inception Distance), and KID (Kernel Inception Distance). $\uparrow$ ($\downarrow$) indicates that higher (lower) values are better. Best results are shown in \textbf{bold}.
}
\label{tab:main_comparison}
\end{table*}

\subsection{Enhanced Positional Encoding for Multi-Resolution Adaptability}
\label{sec:positional_encoding}

Positional encoding is critical in visual transformers for modeling spatial relationships among tokens. While traditional methods \cite{peebles2023scalable, devlin2019bert, zhu2024m} use sinusoidal encodings to represent absolute positions, A‑DiT employs \textbf{Rotary Positional Embedding (RoPE)} \cite{su2024roformer, tan2024ominicontrol}, which jointly captures absolute and relative positional information within a unified formulation. This enhances spatial reasoning and has shown strong performance in large-scale transformer models.

To further enable multiresolution synthesis, we incorporate \textbf{Centralized Interpolative Positional Encoding (CIPE)} \cite{li2024hunyuan}, which aligns positional encodings across feature maps of different resolutions. CIPE maintains consistent spatial semantics during both training and inference, allowing the model to generalize to unseen image sizes. Unlike previous VTO approaches that depend on external pose detectors for alignment, our framework relies solely on RoPE-based encoding, removing additional preprocessing and enabling precise garment placement directly within the transformer.

\subsection{Finetuning for Person Features \& Person Pose}
\label{sec:finetuning_person}


To preserve user-specific identity attributes such as tattoos, scars, muscle, tone and accessories while maintaining garment fidelity, we employ a targeted finetuning strategy within the A-DiT architecture. As described in A-DiT Module section~\ref{sec: A-DiT Module}, person-specific features are processed separately in the diffusion and garment features and remain fixed in the transformer blocks where conditioning is applied. This design allows us to finetune only the person representation rather than the entire diffusion model, reducing computational cost and mitigating overfitting. Empirical results in Experiments Section~\ref{sec: exp} confirm that this selective finetuning preserves garment generalization while maintaining individual identity.


\section{Experiments}
\label{sec: exp}

\subsection{Datasets}
We conducted experiments on two widely-adopted standard virtual try-on datasets, VITON-HD \cite{choi2021viton} and DressCode \cite{morelli2022dress}, which together provide a diverse range of clothing variations and person images for comprehensive evaluation. Since VITON-HD contains only upper garment images, we preprocess this dataset as detailed in Dataset Preparation Section \ref{sec:data_process}. Additionally, because individual images in the DressCode dataset do not consistently include both upper and lower garments, further processing is required to align them with our framework.

\begin{figure*}
    \centering
        \includegraphics[width= 0.95\linewidth]{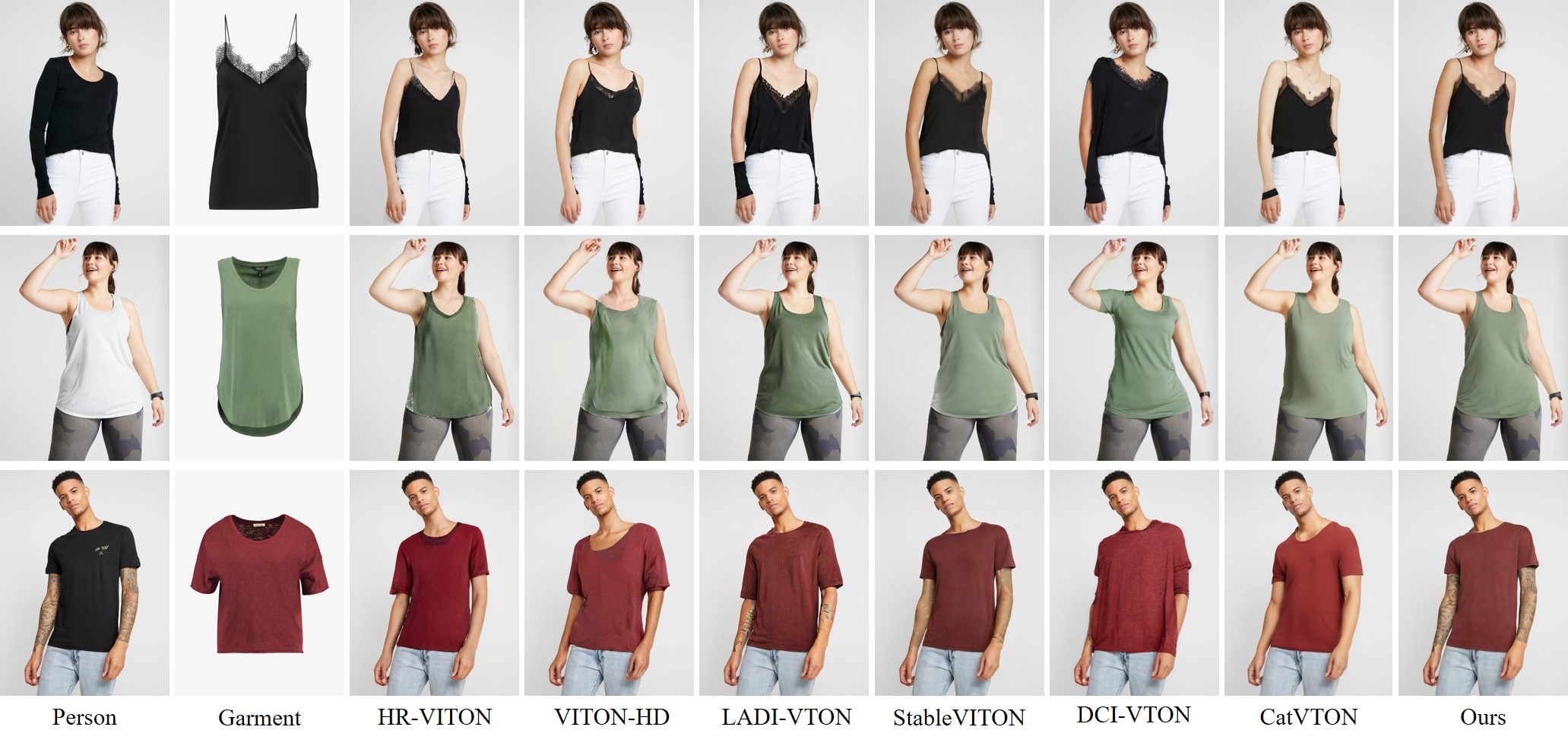}
        \caption{
            \textbf{Qualitative comparison of MEGA‑VTON with existing VTO methods.} 
            Row 1: Only our model preserves both garment and identity features with clear detail. 
            Row 2: MuGa‑VTON generates more realistic garment textures and faithfully retains accessories (e.g., watch). 
            Row 3: Our method uniquely preserves tattoos with high visual fidelity.
        }
        \label{fig:comperison_fig}
        \vspace{-5pt}
\end{figure*}

\subsection{Implementation Details}
\label{sec: implementation_details}

We adopt a two-stage training strategy for our model. In the first stage, the model is trained on $512 \times 256$ images for $800K$ iterations. In the second stage, we initialize the model from the pretrained checkpoint obtained in stage one and finetune the person features for an additional $400K$ iterations. For both stages, we use a batch size of $128$, and the learning rate is linearly warmed up from $10^{-6}$ to $10^{-4}$ over the first $20K$ steps before being maintained at a constant value. We parameterize the model output in $v$-space as described in \cite{salimans2022progressive}, while the L2 loss is computed in $\epsilon$-space. Furthermore, to implement classifier-free guidance \cite{ho2022classifier}, conditional inputs are set to zero in 10\% of the training iterations. 

\subsection{Quantitative Results}
\label{sec: quantitative_results}


We compare MuGa-VTON against nine open-source methods: VITON-HD, HR-VITON, GP-VTON, LADI-VTON, DCI-VTON, StableVITON, CatVTON, IDM-VTON, and Any2AnyTryon on the VITON-HD \cite{choi2021viton} and DressCode-Upper \cite{morelli2022dress} datasets (Table~\ref{tab:main_comparison}). Because most prior work evaluates only upper-garment transfer, we adopt the same setting for fairness; LADI-VTON, the sole multi-garment baseline, is further examined in our user study. Results are reported with four widely used metrics: SSIM \cite{wang2004image}, LPIPS \cite{zhang2018unreasonable}, FID \cite{heusel2017gans}, and KID \cite{binkowski2018demystifying}. MuGa-VTON attains the best SSIM and LPIPS scores on both datasets while matching or surpassing the strongest baselines in FID and KID. In addition, MuGa-VTON uniquely supports prompt-based garment customization, for which no public VTO baseline is currently available.

\begin{table}[htb]
\centering
\resizebox{0.48\textwidth}{!}{%
\begin{tabular}{l|cc|cc}  
\toprule
\textbf{Method} & \multicolumn{2}{c|}{\textbf{Paired Test}}
                & \multicolumn{2}{c}{\textbf{Unpaired Test}} \\
\cmidrule(r){2-3} \cmidrule(l){4-5}  
                & VITON-HD & DressCode & VITON-HD & DressCode \\ 
\midrule
VITON-HD    &  5.69\% &   --    &  3.57\% &   --    \\
HR-VITON    & 13.88\% & 10.48\% & 12.86\% & 10.37\% \\
LADI-VTON   &  6.05\% &  5.71\% & 10.00\% &  9.63\% \\
StableVITON & 19.57\% & 20.95\% & 19.64\% & 18.37\% \\
DCI-VTON    & 10.32\% & 12.38\% &  9.29\% & 13.33\% \\
CatVTON     & 20.28\% & 22.86\% & 21.07\% & 22.85\% \\
\textbf{Ours}&\textbf{24.20\%}&\textbf{27.62\%}&\textbf{23.57\%}&\textbf{25.44\%}\\
\bottomrule
\end{tabular}}
\caption{User-study preference scores (\%); higher is better.}
\label{tab:user_study_table}
\vspace{-5 pt}
\end{table}

\subsection{Qualitative Results}
\label{sec:qualitative_results}

Figure \ref{fig:comperison_fig} shows a visual comparison between MuGa-VTON and several baseline methods. Our approach excels in preserving intricate garment details while accurately maintaining the distinct features of the person. The generated images exhibit remarkable consistency in texture, pattern, and overall fit. Furthermore, as illustrated in Figure \ref{fig:img_editing}, our method demonstrates a superior ability to interpret garment layout cues, enabling precise modifications to targeted areas without inadvertently affecting other regions.

\subsection{User Study}
\label{sec:user_study}

To evaluate human perception, we conducted a user study comprising two evaluation protocols:  
(i) \textbf{Paired Test} – Participants were presented with images where the generated clothing was identical to the reference garment and asked to select the image that best matched the original.  
(ii) \textbf{Unpaired Test} – Participants were shown images where the generated clothing differed from the reference garment and instructed to choose the most visually realistic image.  

In both cases, participants were also required to sort the images in order of perceived realism, with scores assigned based on the resulting rankings. Each experiment was conducted on a set of 100 randomly selected test samples, evaluated by 20 volunteers on Amazon Mechanical Turk (AMT). As shown in Table \ref{tab:user_study_table}, our approach consistently outperformed existing methods in both evaluation settings.

\begin{figure}[ht]
    \centering
    \resizebox{0.48\textwidth}{!}{
    \subfloat[FID Score]{%
        \includegraphics[width=0.48\linewidth]{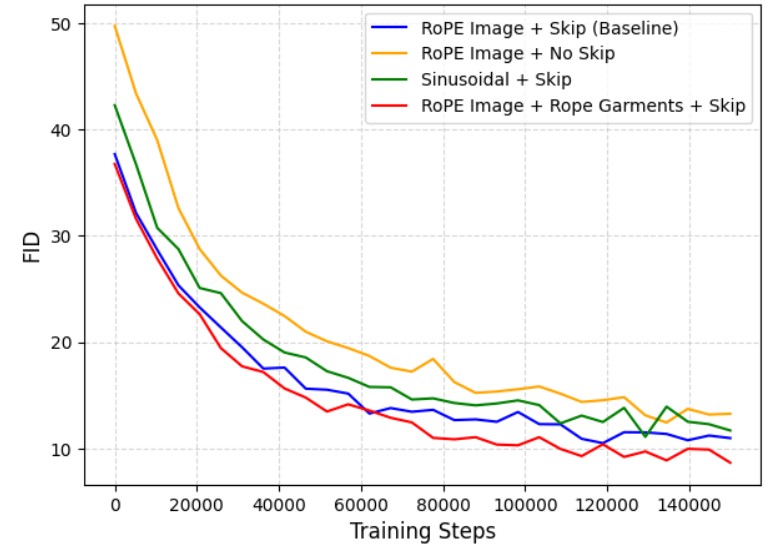}%
        \label{fig:image1}%
    }
    \hfill
    \subfloat[KID Score]{%
        \includegraphics[width=0.48\linewidth]{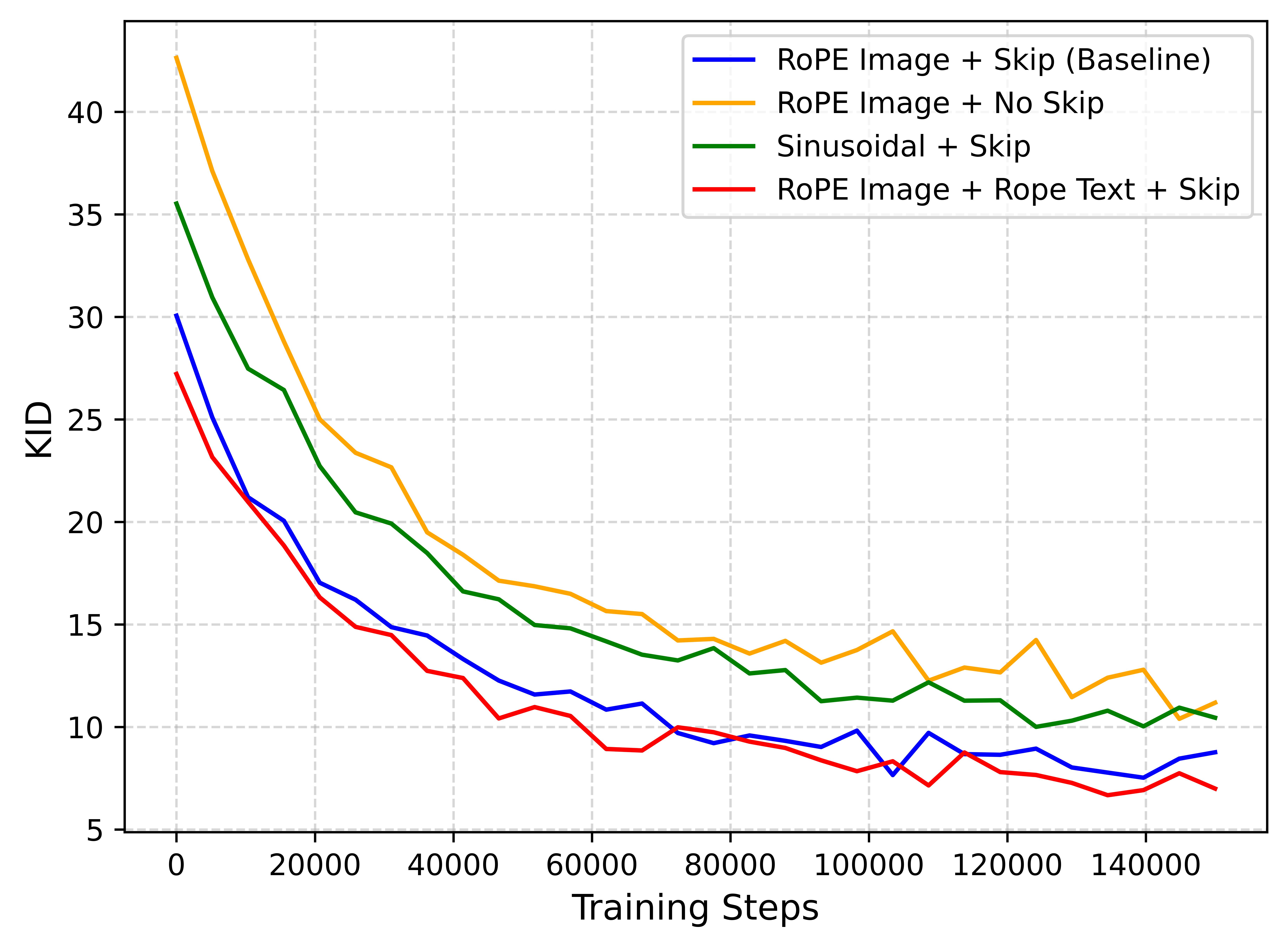}%
        \label{fig:image2}%
    }
    }
    \caption{ Ablation study on position encoding \& skip module.}
    \vspace{-5 pt}
    \label{fig:aba_fig}
    
\end{figure}

\subsection{Ablation Study}
\label{sec:ablation}
In this section, we present an ablation study to demonstrate how key design choices in our approach impact overall performance and reveal potential trade-offs. Specifically, we examine the influence of skip modules, the effectiveness of different positional encoding schemes, and the significance of person pose information in producing realistic try-on results.

\textbf{Impact of the Skip Module} – Skip connections play a crucial role in fusing features between corresponding encoder and decoder layers in U-Net architectures. In our A-DiT model, similar skip modules are employed to enhance information flow. As shown in Figure \ref{fig:aba_fig}, removing these skip connections leads to a significant degradation in performance, with both FID and KID scores increasing noticeably.

\textbf{Comparison of Positional Encoding Schemes} – We investigate the effect of two positional encoding methods: the traditional baseline sinusoidal encoding (used in the original DiT \cite{peebles2023scalable}) and Rotary Positional Embedding (RoPE) \cite{su2024roformer}. The results, presented in Figure \ref{fig:aba_fig}, demonstrate that RoPE consistently outperforms the sinusoidal approach throughout the training process. Moreover, RoPE accelerates model convergence, likely due to its capability to capture both absolute and relative positional information more effectively.

\begin{figure}[t]
    \centering
    \includegraphics[width=0.85\linewidth]{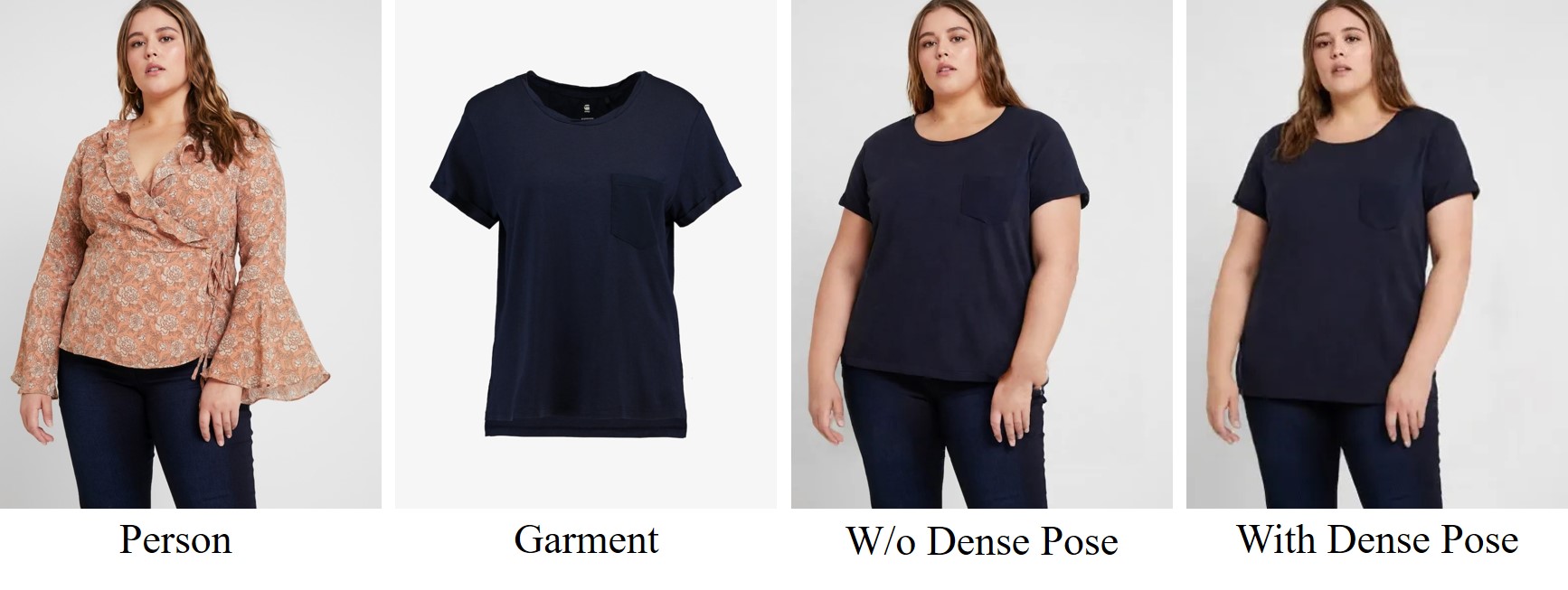}
    \caption{
        Ablation results comparing the impact of including and excluding the pose map estimation during generation.
    }
    \label{fig: abalation_result}
    \vspace{-5pt}
\end{figure}

\textbf{Without Person Pose} – Incorporating a dense pose estimation map (as depicted in Figure~\ref{fig: abalation_result}) is essential for accurately reconstructing body structures. When the pose information is removed, especially for baggy garments, the model sometimes produces distorted arm shapes or misaligned body proportions, underscoring the importance of pose cues for generating realistic try-on results.

\begin{figure}[t]
    \centering
    \includegraphics[width=0.83\linewidth]{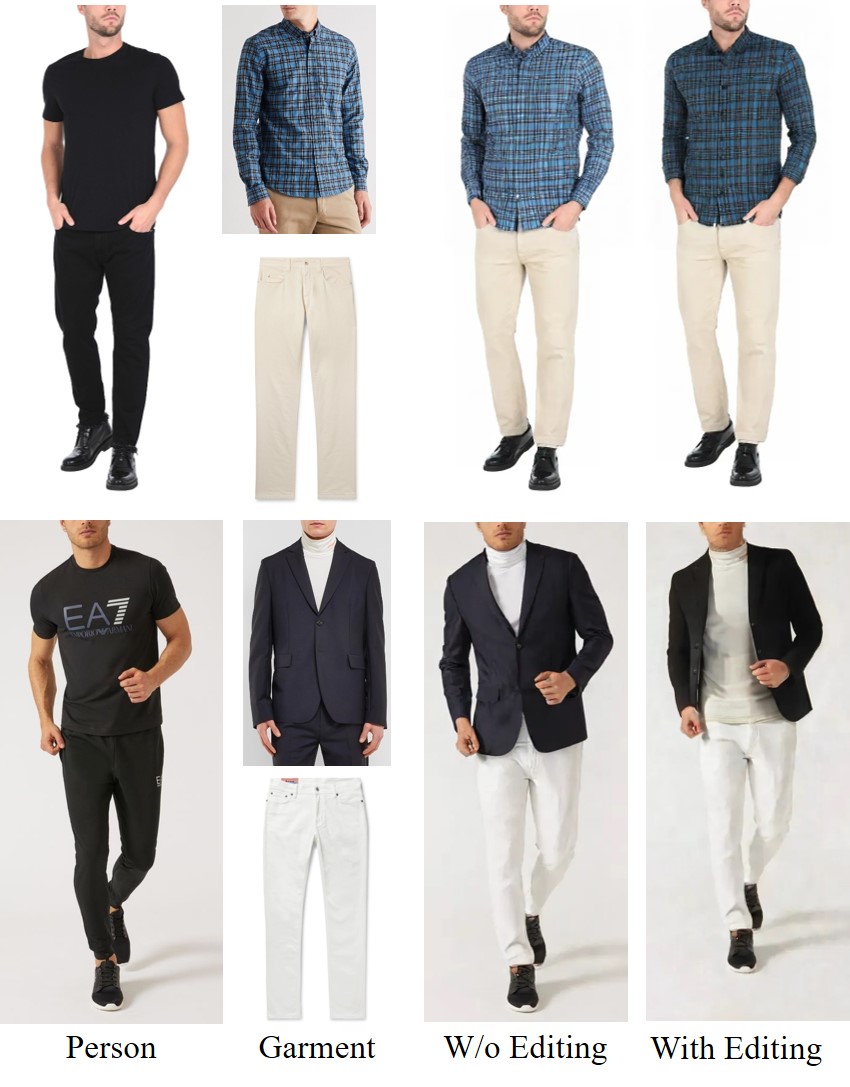}
    \caption{
        Visualization of prompt-based image editing. The first result corresponds to the prompt ``roll up the shirt'' and the second to ``open the outer top''. Each pair shows the generated image with and without the specified prompt.
    }
    \label{fig:img_editing}
    \vspace{-5 pt}
\end{figure}


\subsubsection{Garment Editing}
We introduce a prompt-based garment editing method that enables fine-grained control over garment appearance in generated images. As shown in Figure \ref{fig:img_editing}, prompts like “roll up the shirt” or “open the outer top” are effectively interpreted and applied by the model, while the absence of prompts preserves original garment attributes. This demonstrates the model's ability to selectively modify or retain features. Due to the absence of publicly available prompt-driven editing baselines, direct comparison with existing methods is not currently possible.


\subsubsection{Limitations}
\label{sec:limitations}

\begin{figure}
    \centering
        \includegraphics[width=0.9\linewidth]{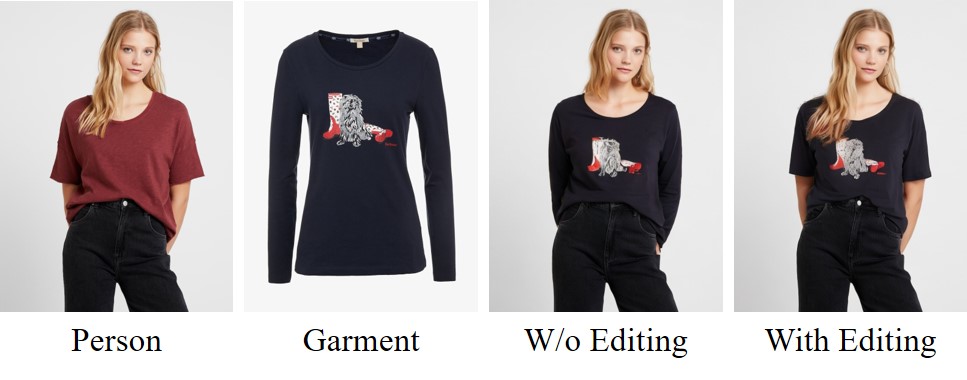}
        \caption{\textbf{Drawback results:} Our model generates a short-sleeve t-shirt when the prompt ``Sleeve Length: short'' is given for a long-sleeve t-shirt (Note: the prompt does not include ``roll up the t-shirt''). This suggests that the model may sometimes hallucinate. Additionally, very complex logos may not be rendered clearly.}
        \label{fig:drawback_result_1}
        \vspace{-5pt}
\end{figure}

While effective, our proposed solution has some constraints. It struggles with detailed garment edits as shown in Figure \ref{fig:drawback_result_1},  e.g., converting long to short sleeves lacks nuanced effects like roll-ups. It also fails to preserve fine fabric textures and clear logos, though fine-tuning on high-res data \cite{zhu2024m} may help. Lastly, the model finds uncommon clothing combinations challenging.


\section{Conclusion}
\label{sec: conclusion}


In this work, we proposed \emph{MuGa‑VTON}, a \emph{Mu}lti-\emph{Ga}rment virtual try-on framework powered by three novel modules: \emph{PRM} (Person Representation Module), \emph{GRM} (Garment Representation Module) and \emph{A‑DiT} diffusion transformer, which jointly model upper and lower garments in a shared latent space and enable prompt-based customization as a key capability. This design allows users to control garment appearance and layout through natural language instructions while preserving identity-specific details. By integrating advanced positional encoding, progressive training, and person-specific feature conditioning, our approach achieves fine-grained garment detail preservation and identity consistency across multiple resolutions. Extensive experiments on VITON‑HD and DressCode, supported by ablation studies and user evaluations, demonstrate that MuGa‑VTON delivers state-of-the-art performance in both qualitative realism and quantitative fidelity.

While the model significantly advances personalization and controllability in virtual try-on, challenges remain in handling rare garment combinations and achieving precise layout control. Future work will explore higher-resolution training, richer prompt interactions, and improved feature fusion to further enhance the practicality and versatility of multi-garment virtual try-on systems.


\bibliography{aaai2026}

\clearpage

\section{Dataset Preparation and Preprocessing}
\vspace{1em}
\hrule
\vspace{1em}

In addition to the details provided in the main paper, here we describe the dataset preparation pipeline and masking procedure in greater depth. Our preprocessing incorporates both layflat product images and cropped garment images to enhance robustness across different input formats. For the DressCode dataset, we extend the standard pipeline to include both upper and lower garment categories, enabling multi-garment try-on.

To generate precise garment masks, we employ \textbf{Sapiens} model, a state-of-the-art human body-part segmentation model capable of segmenting 28 anatomical regions even under challenging poses or occlusions. The high-quality segmentation masks produced by Sapiens allow accurate garment isolation and serve as the foundation for constructing refined agnostic masks used during training.

The masking pipeline proceeds as follows:
\begin{enumerate}
    \item A binary segmentation mask is first generated using Sapiens predictions.
    \item A morphological dilation operation with kernel size $3$ is applied to slightly expand the mask, ensuring coverage of minor misalignments or boundary errors.
    \item A thresholding step is applied to convert the dilated mask into a strictly binary form (0 or 255), removing intermediate gray values and ensuring sharp, artifact‑free garment boundaries for precise extraction.
\end{enumerate}

This refined masking not only isolates the garment accurately but also preserves fine-grained, identity-specific attributes such as tattoos, scars, muscle tone, and accessories. Retaining these high-frequency details is essential for realistic and personalized garment transfer in the proposed MuGa-VTON framework.

\subsection{Prompt Structure for Attribute Annotation}

To generate consistent clothing attribute annotations from images, we employed a structured prompt specifically designed for multimodal models (e.g., LLaVA, GPT-4o). The prompt ensures comprehensive coverage of garment type, fit, and key details (e.g., sleeve length, outerwear, tuck-in status) while enforcing logical consistency.

\begin{figure}
    \centering
    \includegraphics[width=0.99\linewidth]{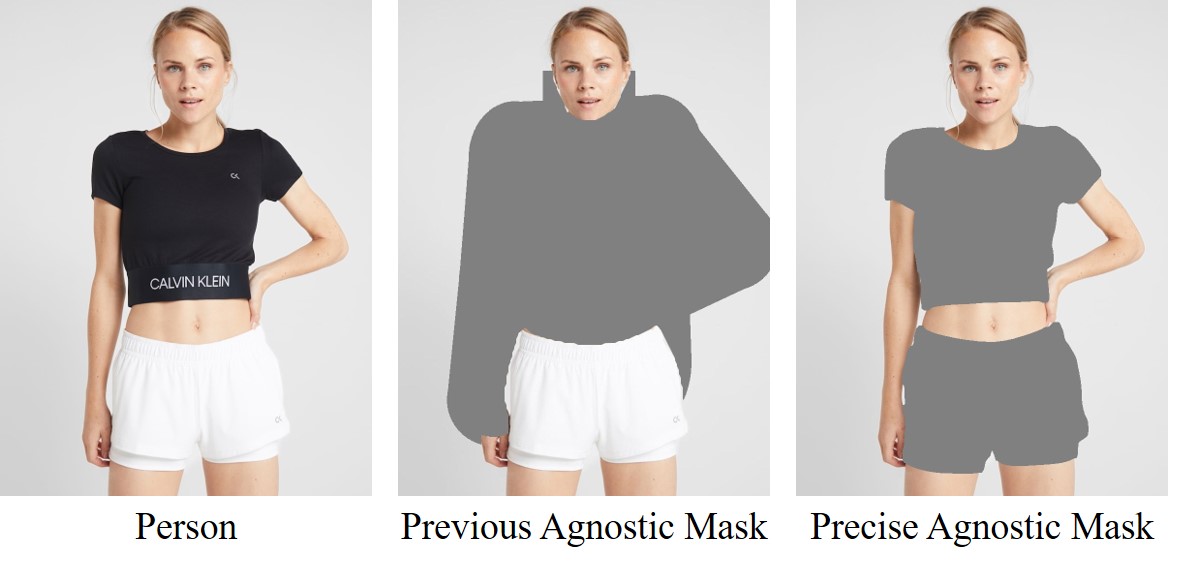}
    \caption{An example of the mask generation process. The figure illustrates how the initial binary mask is refined through dilation and thresholding to accurately capture garment boundaries while preserving essential details of the user's appearance.}
    \label{fig:mask}
\end{figure}

\begin{tcolorbox}[
    title=Prompt Template,
    colback=gray!5,
    colframe=gray!70,
    breakable,
    listing only,
    listing options={
        basicstyle=\ttfamily\scriptsize,
        columns=fullflexible,      
        keepspaces=true,           
        showstringspaces=false
    }
]
\textbf{Do not reference any previous chats or context!}
Carefully analyze the given image. Pay close attention to the clothing details
and provide clear, consistent answers to the following attributes and focus on Guidelines:

- \textbf{Sleeve Length of the Upper Cloth}: (Options: short, long, sleeveless)  \\
- \textbf{Sleeves Rolled Up}: (Options: Yes, No)  \\
- \textbf{Top Tucked In}: (Options: Yes, No)  \\
- \textbf{Wearing Outer Top}: (Options: Yes, No)  \\
- \textbf{Outer Top Open}: (Options: Yes, No)  \\
- \textbf{Fit}: (Examples: tight, loose, regular)  \\
- \textbf{Image Path}: (path of the input image)  \\

\textbf{Guidelines:}
1. Ensure answers are logically consistent (e.g., sleeveless garments cannot have rolled-up sleeves).
2. If uncertain about an attribute, respond with "unknown" rather than guessing.

\textbf{Output:}
Only provide confident attribute values or return "unknown" where certainty is lacking.
\end{tcolorbox}

\paragraph{Usage}
This prompt was passed to the multimodal model along with the input image path. 
Responses were collected via streaming API calls to enable incremental output logging, 
and the final aggregated response was stored as structured annotations for downstream dataset preparation.

\section{Supplementary Results}
Below are some additional results that are not included in the main paper. 

\begin{figure*}
    \centering
        \includegraphics[width=0.99\linewidth]{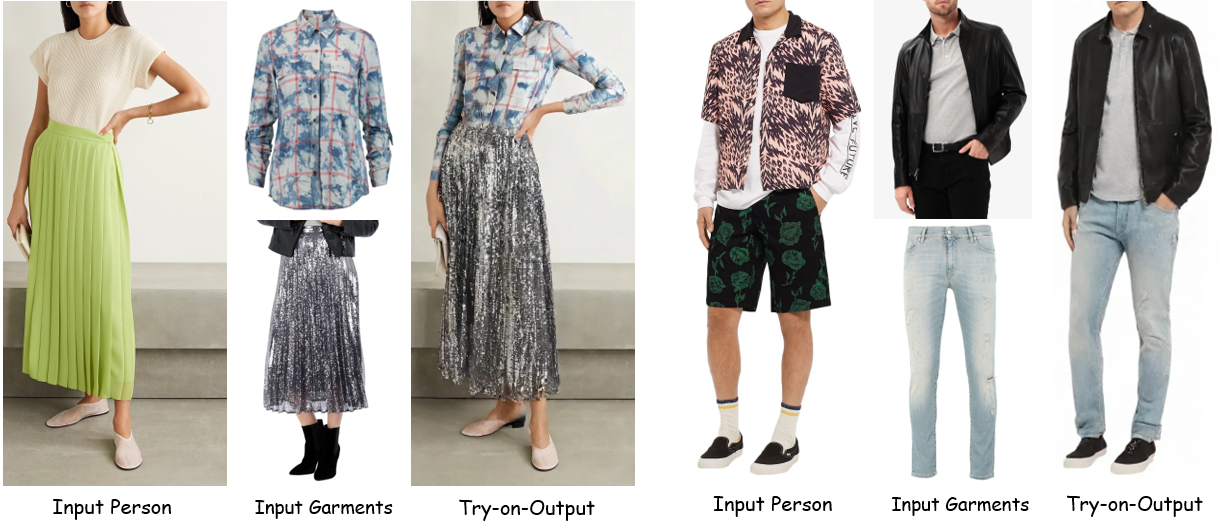}

        \vspace{10 pt}
        
        \includegraphics[width=0.99\linewidth]{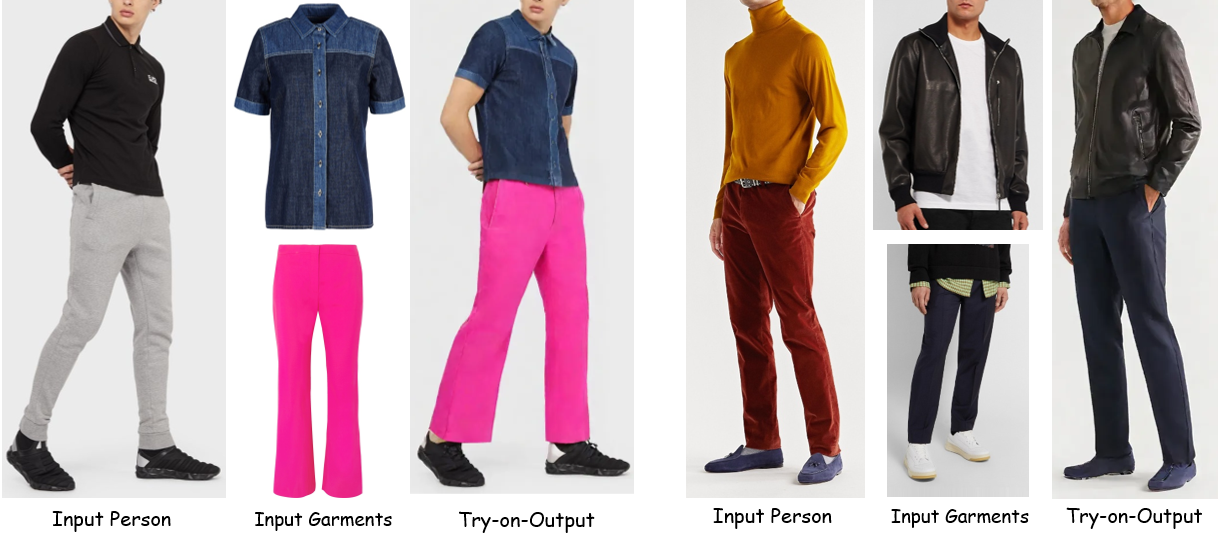}

        \vspace{10 pt}

        \includegraphics[width=0.99\linewidth]{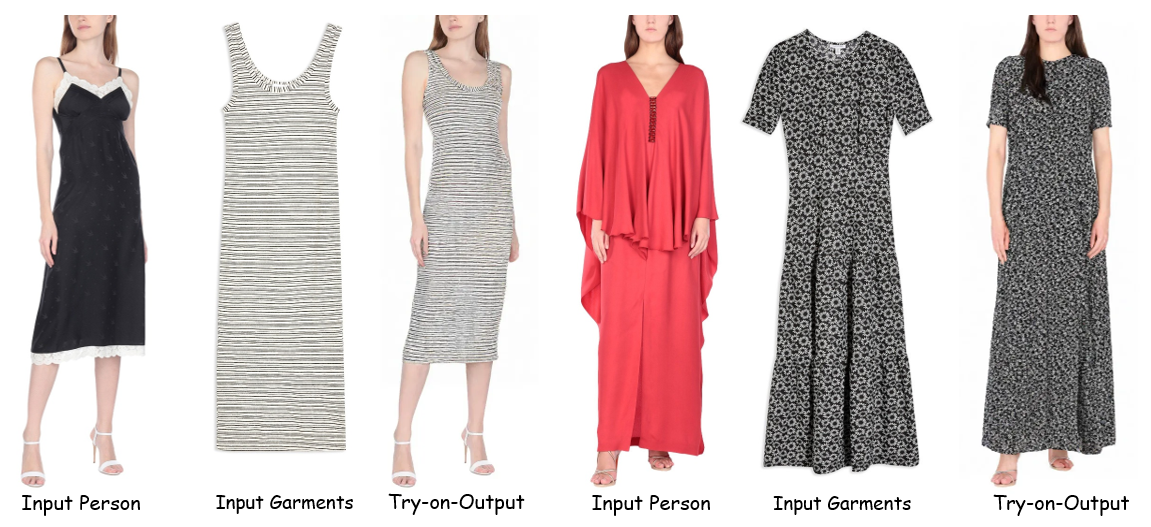}
        \label{fig:supp_result_1}
\end{figure*}

\begin{figure*}
    \centering
        \includegraphics[width=0.99\linewidth]{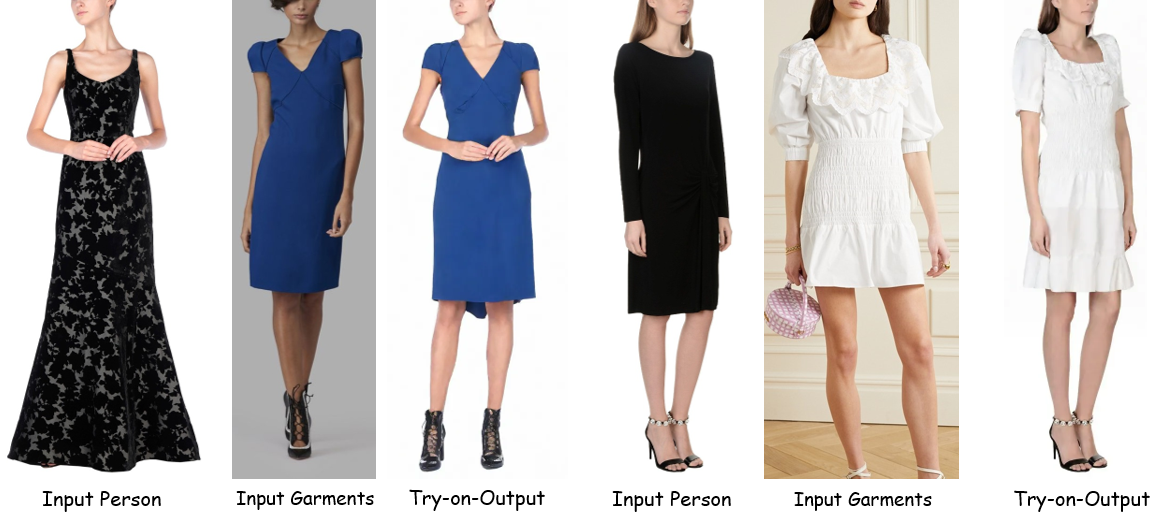}

        \vspace{10 pt}

        \includegraphics[width=0.99\linewidth]{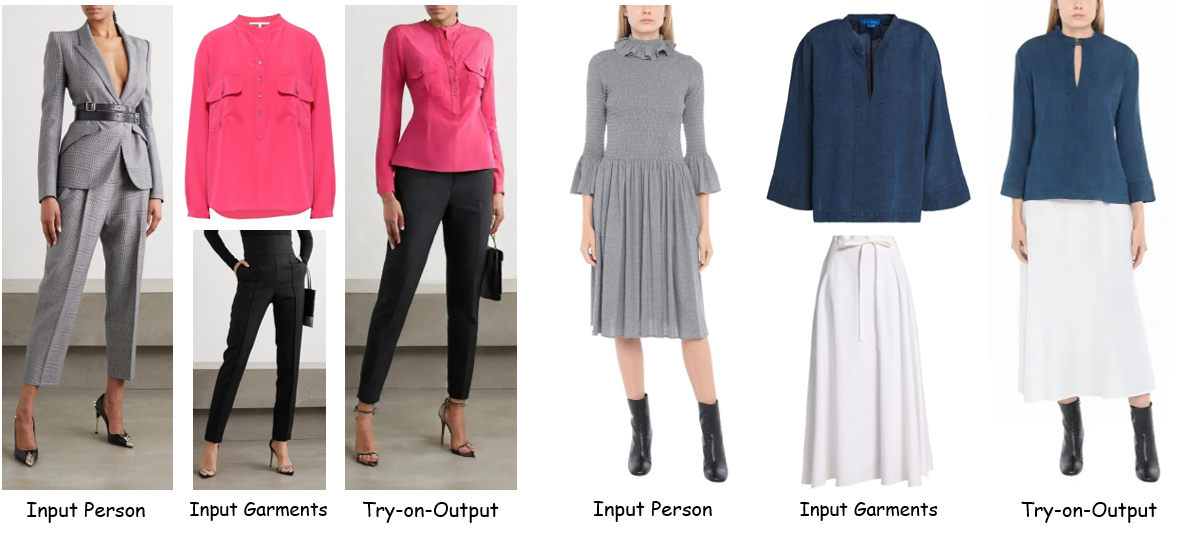}

        \vspace{10 pt}

        \includegraphics[width=0.99\linewidth]{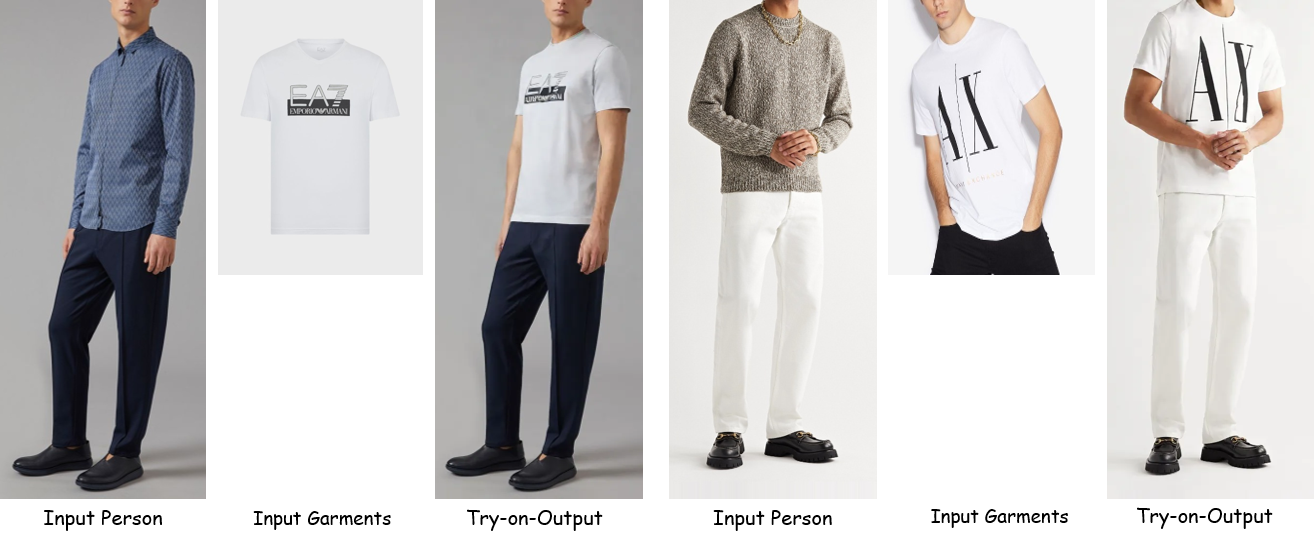}
        \label{fig:supp_result_2}
\end{figure*}

\begin{figure*}
    \centering
        \includegraphics[width=0.99\linewidth]{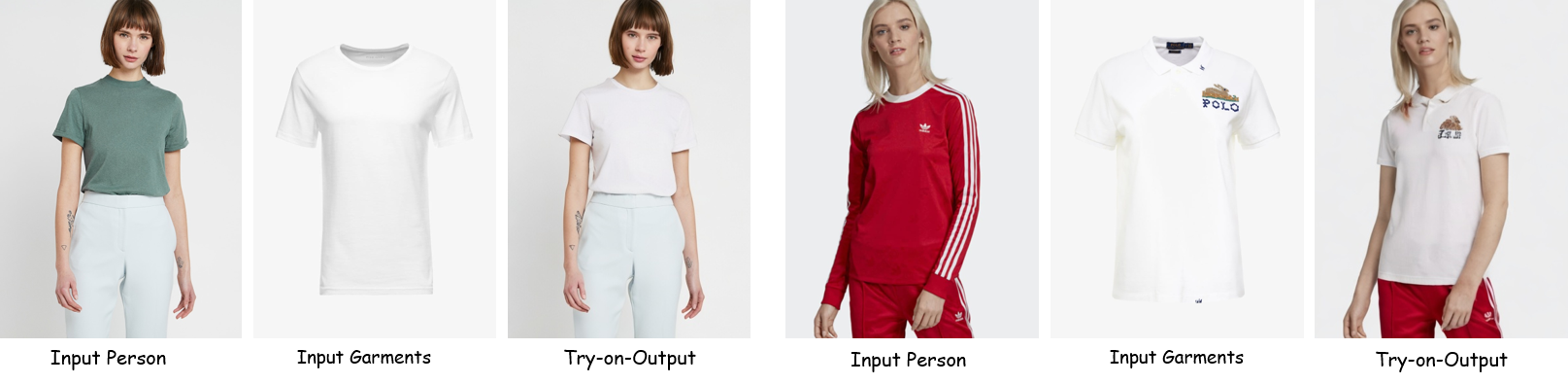}

        \vspace{10 pt}

        \includegraphics[width=0.99\linewidth]{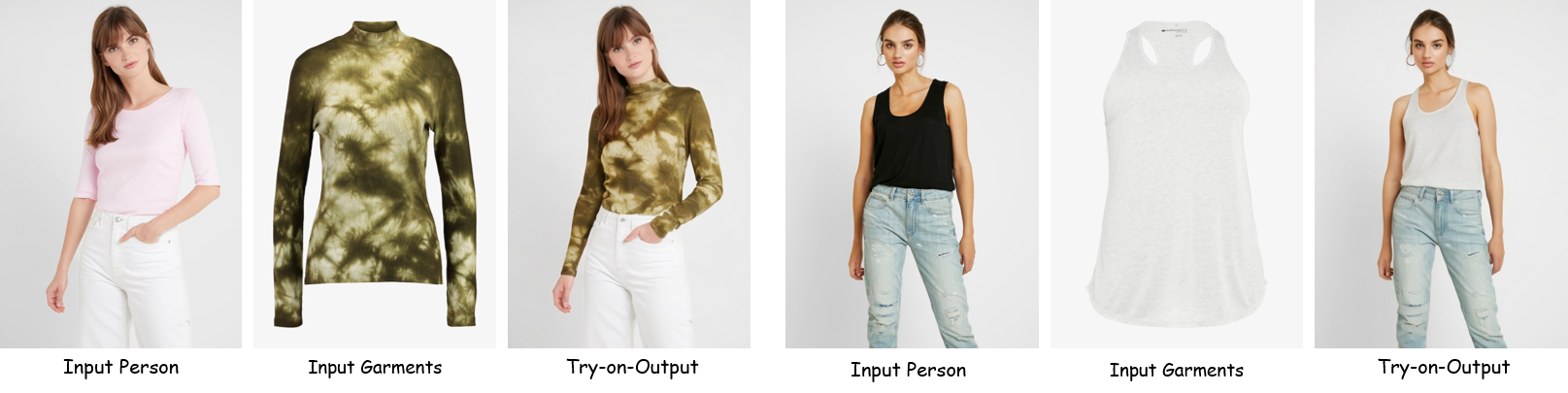}

        \vspace{10 pt}

        \includegraphics[width=0.99\linewidth]{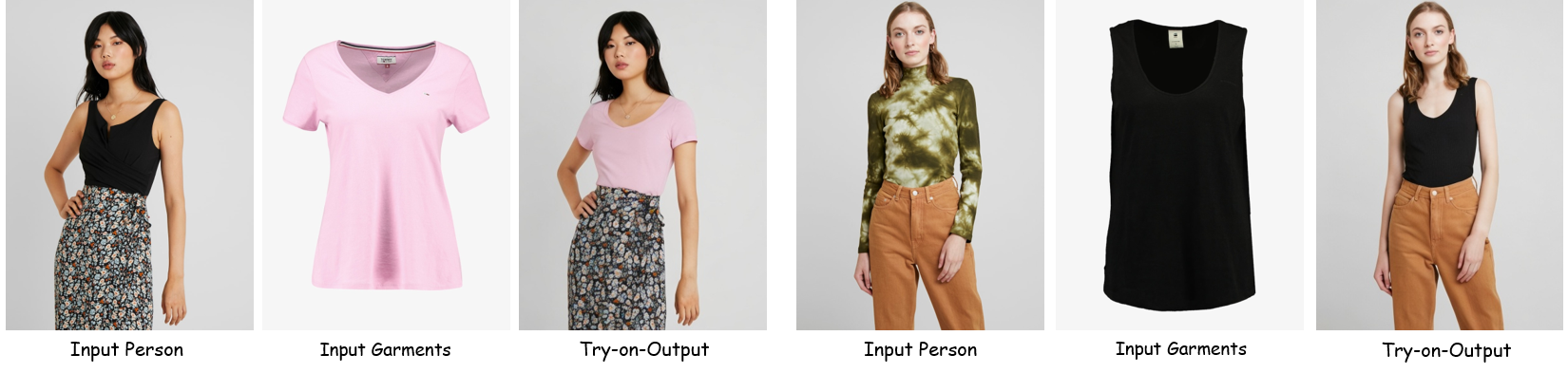}

        \vspace{10 pt}

        \includegraphics[width=0.99\linewidth]{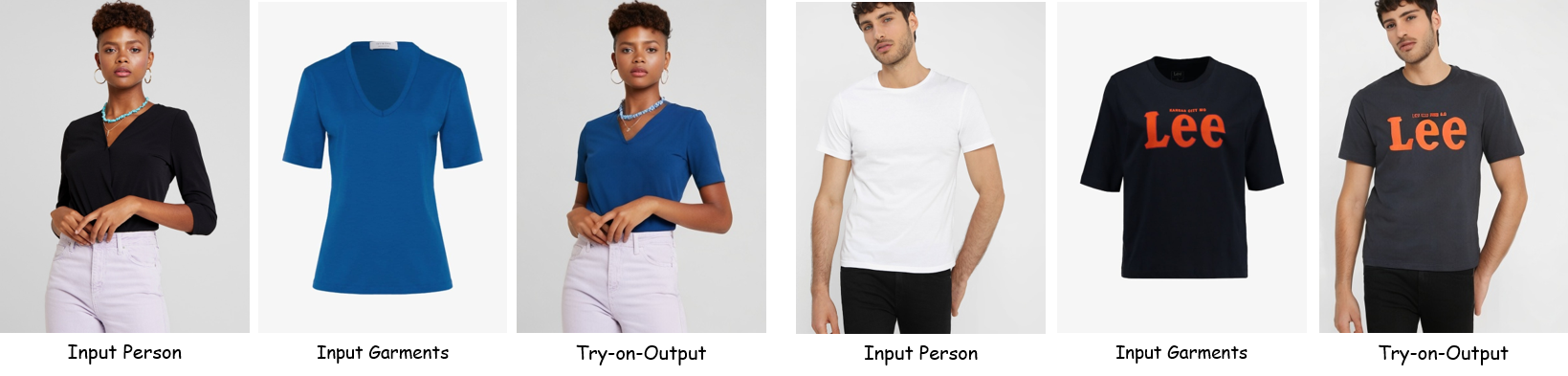}
        \caption{Additional qualitative results generated by \textbf{MuGa-VTON}.  
            Each triplet shows (left) the target person, (middle) the input garments(upper, lower, full), and (right) the synthesized try-on image.  
            The examples span diverse poses, lighting conditions, and garment styles, demonstrating the framework’s ability to preserve identity-specific details (e.g., tattoos, accessories) while accurately rendering multiple garments and supporting prompt-based edits.
        }

        \label{fig:supp_result_3}
\end{figure*}


\end{document}